\documentclass{article}

\usepackage{microtype}
\usepackage{graphicx}
\usepackage{booktabs} 

\PassOptionsToPackage{hyphens}{url} 
\usepackage{hyperref}
\usepackage{tikz}
\usepackage{enumitem}

\usepackage[accepted]{icml2024}

\usepackage{amsmath}
\usepackage{amssymb}
\usepackage{mathtools}
\usepackage{amsthm}

\usepackage[capitalize,noabbrev]{cleveref}

\theoremstyle{plain}

\theoremstyle{definition}

\theoremstyle{remark}

\usepackage[textsize=tiny]{todonotes}
\newcommand{\parencite}[1]{\cite{#1}}
\newcommand{\textcite}[1]{\citet{#1}}

\usepackage{xspace}
\usepackage{subcaption}
\usepackage{placeins}
\icmltitlerunning{Uncertainty Quantification for Neural Weather Models}

\begin{document}

\twocolumn[
\icmltitle{Valid Error Bars for Neural Weather Models using Conformal Prediction}



\icmlsetsymbol{equal}{*}

\begin{icmlauthorlist}
\icmlauthor{Vignesh Gopakumar}{equal,ucl,ukaea}
\icmlauthor{Joel Oskarrson}{equal,link}
\icmlauthor{Ander Gray}{equal,ukaea}
\icmlauthor{Lorenzo Zanisi}{ukaea}
\icmlauthor{Stanislas Pamela}{ukaea}
\icmlauthor{Daniel Giles}{ucl}
\icmlauthor{Matt Kusner}{ucl}
\icmlauthor{Marc Deisenroth}{ucl}

\end{icmlauthorlist}

\icmlaffiliation{ucl}{Department of Computer Science, University College London, London, United Kingdom}
\icmlaffiliation{ukaea}{Computing Division, UK Atomic Energy Authority, Oxford, United Kingdom}
\icmlaffiliation{link}{Department of Computer and Information Science, Linköping University, Linköping, Sweden}

\icmlcorrespondingauthor{Vignesh Gopakumar}{v.gopakumar@ucl.ac.uk}
\icmlcorrespondingauthor{Joel Oskarsson}{joel.oskarsson@liu.se}

\icmlkeywords{Surrogate Models, Weather, Climate, Uncertainty Quantification, Conformal Prediction}

\vskip 0.3in
]



\printAffiliationsAndNotice{\icmlEqualContribution} 
           
\begin{abstract}
Neural weather models have shown immense potential as inexpensive and accurate alternatives to physics-based models. 
However, most models trained to perform weather forecasting do not quantify the uncertainty associated with their forecasts.
This limits the trust in the model and the usefulness of the forecasts.
In this work we construct and formalise a conformal prediction framework as a post-processing method for estimating this uncertainty. 
The method is model-agnostic and gives calibrated error bounds for all variables, lead times and spatial locations.
No modifications are required to the model and the computational cost is negligible compared to model training.
We demonstrate the usefulness of the conformal prediction framework on a limited area neural weather model for the Nordic region. We further explore the advantages of the framework for deterministic and probabilistic models. 
\end{abstract}

\section{Introduction}
Weather and climate models are essential tools for forecasting future changes in temperature, precipitation patterns, and extreme weather events. However, these models are characterised with inherent uncertainty arising from the complexity of the Earth's climate system and the challenges in representing all relevant processes and feedbacks \cite{ensembleforecasting_jcp}. 
By using neural networks for weather and climate modelling, researchers can leverage the power of machine learning to capture complex patterns and relationships in climate data, potentially improving the accuracy of predictions.

Our work relies on quantifying the uncertainty associated with these neural-network-driven weather models by way of Conformal Prediction~(CP). Uncertainty Quantification~(UQ) is crucial for ensuring that forecasts are trustworthy and actionable. By providing robust uncertainty estimates for neural weather models, we can help decision-makers understand the range of possible outcomes with varying levels of confidence in different scenarios. This information is essential for developing effective weather management policies and mitigating adverse weather events \parencite{bodnar2024aurora}.

\section{Related Work}
Traditionally, within numerical weather forecasting, uncertainty estimates are obtained via ensemble methods \parencite{ensembleforecasting_jcp}.
These methods generate ensemble forecasts as samples from the distribution of possible future atmospheric states.
Ensemble forecasting is achieved in physical models by introducing perturbations to the initial conditions \parencite{Buizza2008} and to the model parameterisations \parencite{Palmer2009}.
Also neural weather models largely rely on ensemble methods to provide UQ.
These ensemble are created based on perturbing initial states with noise \cite{pathak2022fourcastnet,Bi2023,neural_ensemble_svd}, using sets of different neural network parameters \cite{calibration_of_large_neurwp} or by training generative models \cite{price2024gencast}.
Common to all ensemble methods is that they require rolling out multiple forecasts in order to produce uncertainty estimate.
As an alternative, post-hoc approaches perform UQ across the model outputs as a post-processing step \parencite{bulte2024uncertainty, H_hlein_2024}. 
Conformal Prediction as a method of adding post-hoc uncertainty estimates has gained popularity in the recent past, finding a range of applications within spatio-temporal modelling \parencite{sun2022conformal, ma2024calibrated, CP_Wildfire}. 

\section{Conformal Prediction}
\label{sec: conformal_prediction}

CP \parencite{shafer2008tutorial} gives an answer to the following question: Given some arbitrary dataset $(X_1,Y_1), (X_2,Y_2), ..., (X_n, Y_n)$, and some machine learning model $\hat{f}$ trained on this dataset, what is the accuracy of $\hat{f}$ at predicting the next true label $Y_{n+1}$ at query point $X_{n+1}$. Conformal prediction extends the point prediction $\hat{y}$ of $\hat{f}$ to a prediction set $\mathbb{C}^{\alpha}$, which is guaranteed to contain the true label $Y_{n+1}$ with probability $1-\alpha$

\begin{equation}
    \label{eq: coverage}
    \mathbb{P}(Y_{n+1}\in \mathbb{C}^{\alpha}) \geq 1 - \alpha.
\end{equation}

CP is attractive for quantifying errors in machine learning since the inequality in \cref{eq: coverage} is guaranteed regardless of the selected machine learning model and the training dataset $\left\{(X_{i}, Y_{i})\right\}_{i=1}^{n}$, other than that the samples utilised for the calibration (estimating model performance) being exchangeable (a weaker form of i.i.d.). 

There have been several variants of CP since its original proposal by \textcite{vovk2005algorithmic}. Inductive conformal prediction \parencite{papadopoulos2008inductive}, which we pursue in this work, splits the training set into a \textit{proper training set} (the usual training set for the underlying ML model) and a \textit{calibration set}, used to construct the prediction sets $\mathbb{C}$. The inductive CP framework further extended in this paper, follows a three-step procedure as outlined in \cref{fig: cp_framework_tikz}. It starts with a calibration, followed by estimation of the quantile from the Cumulative Distribution Function (CDF) of non-conformity scores and use the estimated quantile to obtain the prediction sets. 

\tikzstyle{process} = [rectangle, minimum width=0.6cm, minimum height=1.cm, text centered, draw=black]
\tikzstyle{arrow} = [thick,->]

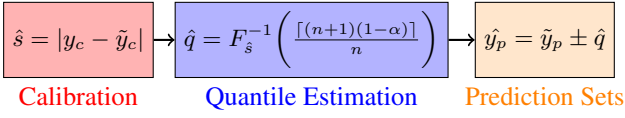
\begin{figure}
    \centering
        \begin{tikzpicture}[remember picture, node distance=3.1cm, font=\small]
            \node (start) [process, fill=red!30, label={[font=\color{red}]below:Calibration}] {$\hat{s} = |y_c - \tilde y_c|$};
            \node (step) [process, fill=blue!30, label={[font=\color{blue}]below:Quantile Estimation}, right of=start] {$\hat{q} = F^{-1}_{\hat{s}}\bigg(\frac{\lceil(n+1)(1-\alpha)\rceil}{n}\bigg)$};
            \node (end) [process, fill=orange!20, label={[font=\color{orange}]below:Prediction Sets}, right of=step] {$\hat{y_p} = \tilde y_{p} \pm \hat{q}$};
        
            \draw [arrow] (start) -- (step);
            \draw [arrow] (step) -- (end);
        
        \end{tikzpicture}
    \caption{Inductive CP Framework over a Deterministic Model (see RES in \cref{nonconformity scores}): 
    (1) Perform calibration using a non-conformity metric (L1 error residual with $\hat{s}$ representing the calibration scores, $y_c, \tilde y_c$ the calibration targets and predictions respectively). 
    (2) Estimate the quantile corresponding to the desired coverage from the CDF of the non-conformity scores ($n$ represents the calibration sample size, $(1-\alpha)$ the desired coverage, $F_{\hat{s}}^{-1}$ the quantile function applied over the inverse CDF of non-conformity scores, $\hat{q}$ the quantile matching the desired coverage). 
    (3) Apply the quantile to the model predictions to estimate the prediction sets ($\tilde y_p$, the model predictions and $\hat{q}$ the upper and lower bars for the predictions).
    }
    \label{fig: cp_framework_tikz}
\end{figure}

\subsection{Conformal Prediction over a Spatio-Temporal Domain}

Neural Weather Models are trained to map the evolution of spatio-temporal field variables.
Starting with an initial distribution of the field variables across the spatial domain, the temporal evolution is mapped by the model in an autoregressive manner. Each model output, spanning the spatio-temporal domain of interest, is characterised as one data point within our framework. Exchangeability is maintained as each calibration sample is taken across the entire spatio-temporal domain of the output. Within our framework, we perform calibration for each cell individually, while preserving the tensorial structure.  This allows us to estimate the marginal coverage for each cell, characterising a spatio-temporal point with a specific variable of interest. CP is performed across each spatio-temporal point output by the model, resulting in upper and lower coverage bands for each point. Upon calibrating for the error bars cell-wise, for validation, we average over all to estimate the coverage across the simulation domain. Within the estimation of the prediction sets, we don't consider the influence of adjacent field points and implicitly expect the model to extract that within the learning process. We find that this approach maintains exchangeability across the various forecasts, allowing us to do CP across spatio-temporal data efficiently. 

\subsubsection{Formal Definition}
\label{subsec: formal_definition}

We define a model  $\mathcal Y = \hat{f}(\mathcal X)$, that learns to map the evolution of an initial temporal sequence of spatial fields ($\mathcal X \in \mathbb{R}^{T_{\text{in}} \times N_x \times N_y \times N_{var}} $) to a later temporal sequence of spatial fields ($\mathcal Y \in \mathbb{R}^{T_{\text{out}} \times N_x \times N_y \times N_{\text{var}}} $). The model inputs and outputs are characterised by 4D tensors, where $T_{\text{in}}, T_{\text{out}}$ represents the temporal dimension of the initial states and forecast respectively, $N_x$ represents the $x$-dimension, $N_y$ represents the $y$-dimension and $N_{\text{var}}$ the field dimension (number of modelled weather variables). The calibration procedure is defined as $\hat{q} = \hat{C}(\mathcal Y, Y)$, utilising the model prediction ($\mathcal Y$) and ground truth ($Y$) to estimate the quantile ($\hat{q}$) associated with the desired coverage. The operation is executed in a point-wise manner as each $\mathcal Y, Y, \hat{q} \in \mathbb{R}^{T_{\text{out}} \times N_x \times N_y \times N_{\text{var}}}$. The quantile is further utilised (as given in \cref{nonconformity scores}) to obtain the lower ($L$) and upper error bars ($U$) across all the cells form the prediction set $\mathbb{C}$, where $L$ and $U$ have the same dimensionality as $\hat{q}$. At inference for a prediction point $X_{n+1}$ with true label $Y_{n+1}$, the expectation of coverage i.e. the prediction set, is guaranteed to satisfy

\begin{equation} \label{eq:coverage_tensor}
    \mathbb{E}\bigg[ (Y_{n+1} \geq L) \wedge (Y_{n+1} \leq U) \bigg]\geq 1 - \alpha.
\end{equation}

\cref{eq:coverage_tensor} is statistically guaranteed to hold for each cell of the spatio-temporal tensor, provided we provide sufficient samples and exchangeability is maintained \parencite{vovk2012conditional}.

\subsubsection{Non-conformity Scores}
\label{nonconformity scores}

A non-conformity score  can be described as a measure of the model's performance across the calibration dataset. Thus, the scores are expressed as a function of the trained model, the calibration inputs and outputs \parencite{gentle_introduction_CP}. Being expressed as the deviation of the model from the ground truth,  we focus on two methods of estimating the non-conformity scores: 

\begin{itemize}[topsep=2pt, itemsep=1pt]
    \item \textbf{Absolute Error Residual (RES):} Utilised for deterministic models, this method involves using the absolute error of the model across the labelled calibration dataset \parencite{error_residual}. The RES non-conformity score is $s(x,y) = |y - \Tilde{f}(x)|$. Upon deriving the quantile ($\hat q$), the prediction set is obtained as : $\{\Tilde{f}(x) - \hat{q}, \; \Tilde{f}(x) + \hat{q}\}$.
    
    \item \textbf{Standard Deviation (STD):} Utilised for probabilistic models, that output a Gaussian predictive distribution with a mean $(\mathbb \mu(x))$ and standard deviation $(\sigma(x))$. The non-conformity score, $s(x,y) = \frac{y - \mathbb \mu(x)}{\sigma(x)}$, takes in the uncertainty of the model and conformalises it to provide validity over the predictive uncertainty. Upon deriving the quantile ($\hat q$), the prediction set is obtained as : $\{\mu(x) - \hat{q} \sigma(x), \; \mu(x) + \hat{q} \sigma(x)\}$. 
\end{itemize}

\section{Neural Weather Models}

\newcommand{\lammodel}{Hi-LAM\xspace}
\newcommand{\lammodelmse}{\lammodel (MSE)\xspace}
\newcommand{\lammodelnll}{\lammodel (NLL)\xspace}
\newcommand{\armapping}{g}


A multitude of neural network based approaches have been successfully applied to  weather forecasting \cite{lam2022graphcast,Bi2023,pathak2022fourcastnet}. 
Irrespective of the choice of architecture and training conditions, neural weather models operate to take in initial states $\mathcal{X}$ and map to a forecast $\mathcal{Y}$. Within the scope of this paper, we focus on the \lammodel model of \citet{neural_lam}. \lammodel is a graph-based neural weather prediction model \cite{keisler, lam2022graphcast}, where a hierarchical Graph Neural Network (GNN) is utilized for producing the forecast.

Let $X^t$ denote the full weather state at time step $t$, including multiple atmospheric variables modelled for all grid cells in some discretisation.
Examples of such atmospheric variables are temperature, wind, and solar radiation.
The GNN $\armapping$ in \lammodel represents the single time step prediction
\begin{equation}
    \label{eq:forecasting_ar}
    X^{t+1} = \armapping\left(X^{t-1:t}, F^{t+1}\right)
\end{equation}
where $F^{t+1}$ are known forcing inputs that should not be predicted.
\Cref{eq:forecasting_ar} can be applied iteratively to roll out a complete forecast of $T$ time steps.
The full forecasting model can thus be viewed as mapping from initial weather states $\mathcal{X} = X^{-1:0}$ and forcing $F^{1:T}$ to a forecast $\mathcal{Y} = X^{1:T}$ of shape $T_{\text{out}} \times N_x \times N_y \times N_{\text{var}}$.
As \lammodel is a limited area model, it produces weather forecasts for a specific sub-area of the globe.
To achieve this, boundary condition along the edges of the forecasting area are given as forcing inputs.


The \lammodel models used in our experiments were trained on the original dataset of forecasts from the MetCoOp Ensemble Prediction System (MEPS) \cite{meps}.
Forecasts are produced for a limited area covering the Nordic region.
One forecast includes $N_\text{var} = 17$ variables modelled on a $N_x \times N_y = 238 \times 268$ grid over $T = 19$ time steps (up to 57 hour lead time).
We refer to \citet{neural_lam} for further details about the model and data.

We use two versions of \lammodel, trained with different loss functions to build a deterministic and a probabilistic model:
\begin{itemize}[topsep=2pt, itemsep=1pt]
    \item \textbf{\lammodelmse:} The exact model from \citet{neural_lam}, trained with a weighted MSE loss.
    This model outputs only a single prediction, to be interpreted as the mean of the weather state.
    \item \textbf{\lammodelnll:} A version of \lammodel that outputs both the mean and standard deviation for each time, variable and grid cell.
    This model was trained with a Negative Log-Likelihood (NLL) loss, assuming a diagonal Gaussian predictive distribution (also called uncertainty loss \cite{fengwu}).
    Apart from the change of loss function, the training setup was unchanged.
\end{itemize}

We use forecasts from September 2021 as our calibration dataset and forecasts from September 2022 as test data.
By using the same month for calibration and testing we minimize the effect of distributional shifts due to seasonal effects, and can work under the assumption of exchangeability. 
Having access to calibration data from the same month, collected the previous year, is a reasonable assumption in practical settings.
For \lammodelmse we compute non-conformity scores using the RES strategy and for the probabilistic \lammodelnll we use STD scores.

\begin{figure}[tb]
    \centering
    \begin{subfloat}
        \centering
        \includegraphics[width=0.9\linewidth]{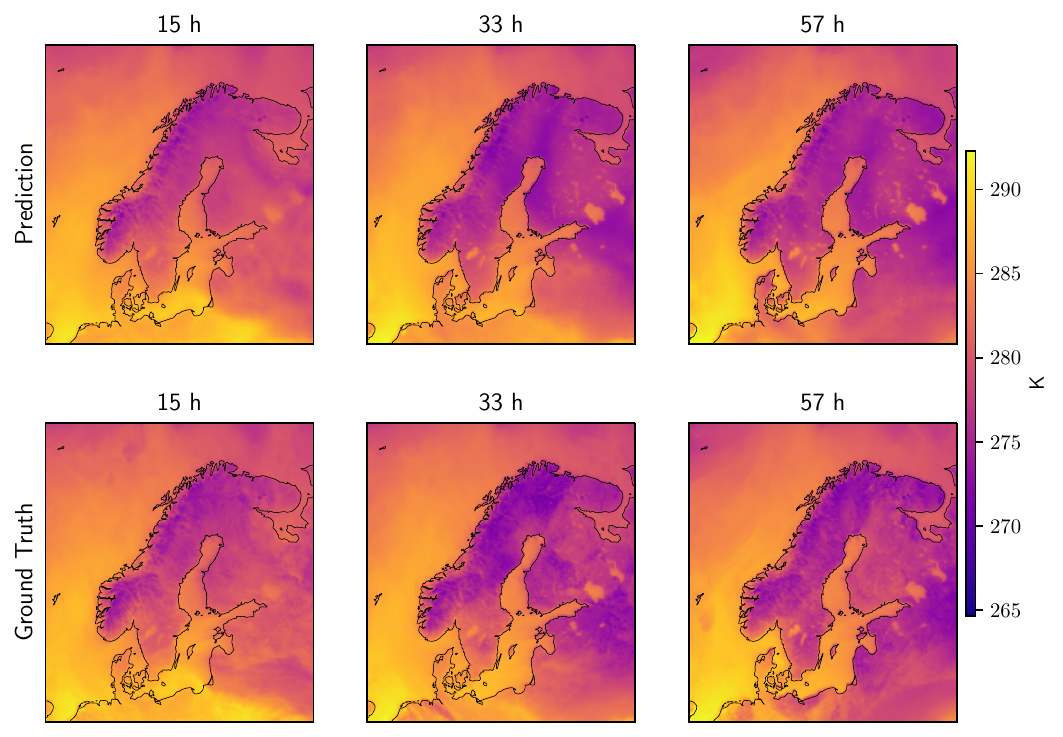}
    \end{subfloat}
    \begin{subfloat}
        \centering
        \hspace{0.03\linewidth}%
        \includegraphics[width=0.87\linewidth]{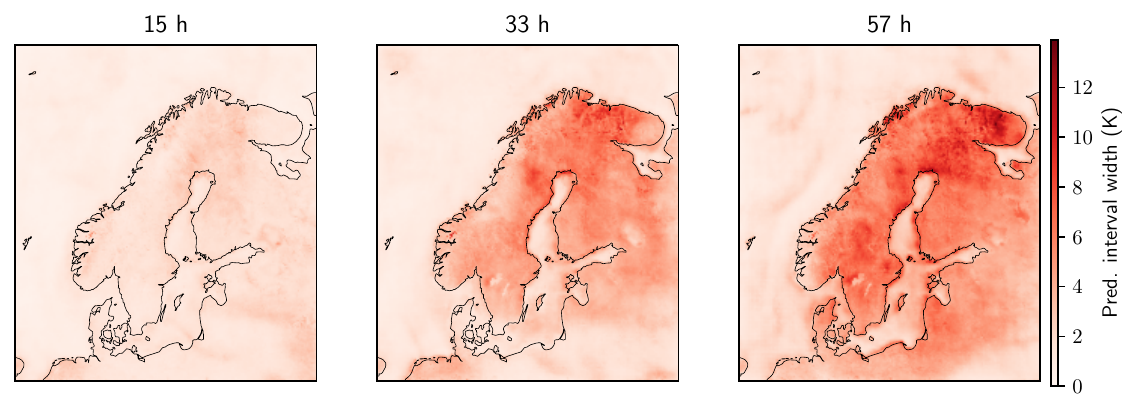}
    \end{subfloat}
    \caption{Prediction (top), Ground Truth (middle) and width of the error bars (bottom) for predicting the temperature 2m above ground (\texttt{2t}) using \lammodelmse.}
    \label{fig:temp_2m}
\end{figure}

\begin{figure}[t] 
    \centering
    \begin{subfloat}
        \centering
        \includegraphics[width=0.9\linewidth]{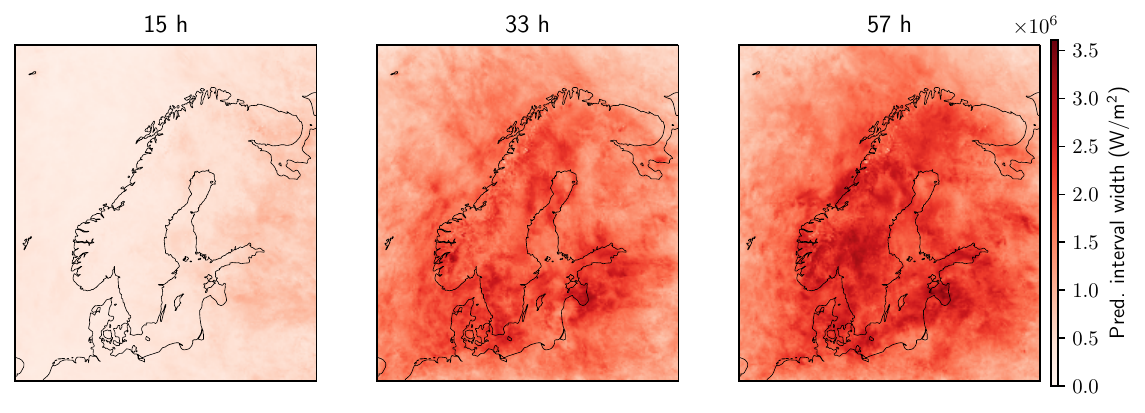}
        \caption*{\lammodelmse}
        \label{fig:neurwp_iwidth_nswrs_mse}
    \end{subfloat}
    \begin{subfloat}
        \centering
        \includegraphics[width=0.9\linewidth]{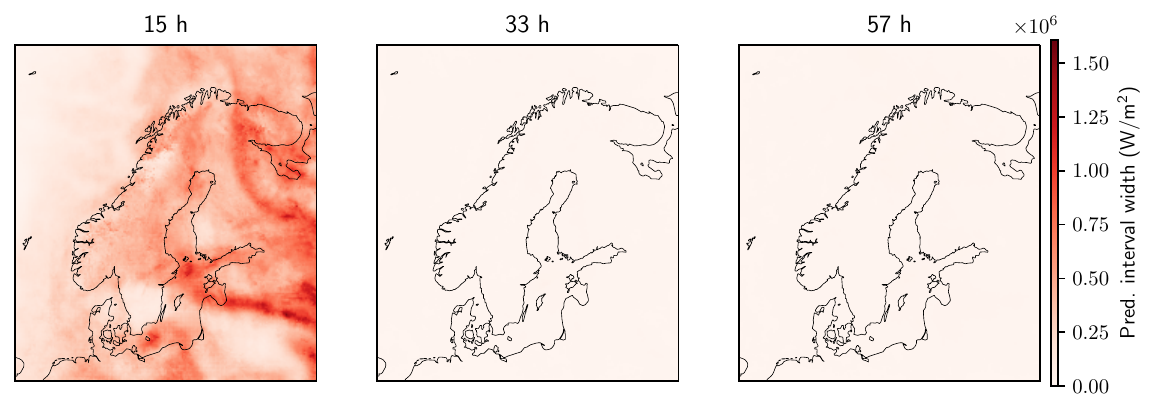}
        \caption*{\lammodelnll}
        \label{fig:neurwp_iwidth_nswrs_nll}
    \end{subfloat}
    \caption{Width of predictive interval at $\alpha = 0.05$ for net shortwave solar radiation flux (\texttt{nswrs}).}
    \label{fig:neurwp_iwidth_nswrs}
\end{figure}

\subsection{Results}

\begin{figure}[t]
    \begin{subfloat}
        \centering
        \includegraphics[width=0.9\linewidth]{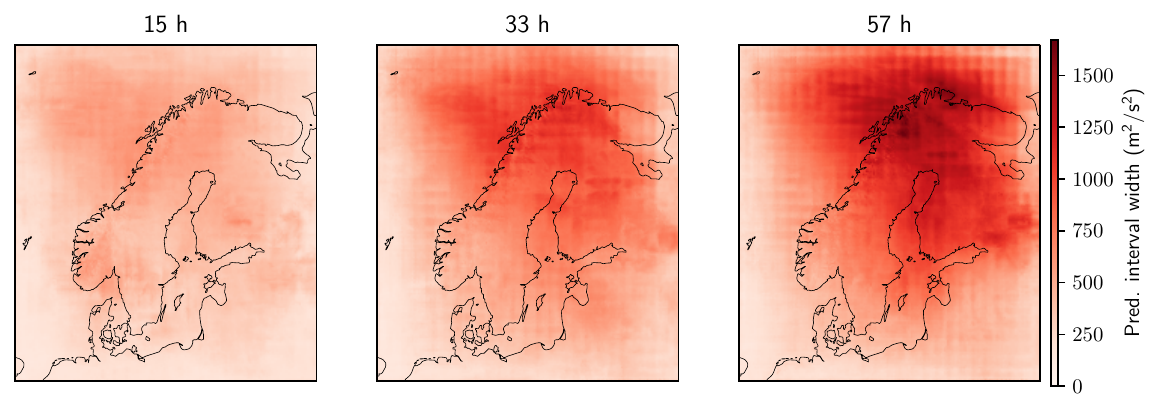}
        \caption*{\lammodelmse}
        \label{fig:neurwp_iwidth_z_mse}
    \end{subfloat}
    \begin{subfloat}
        \centering
        \includegraphics[width=0.9\linewidth]{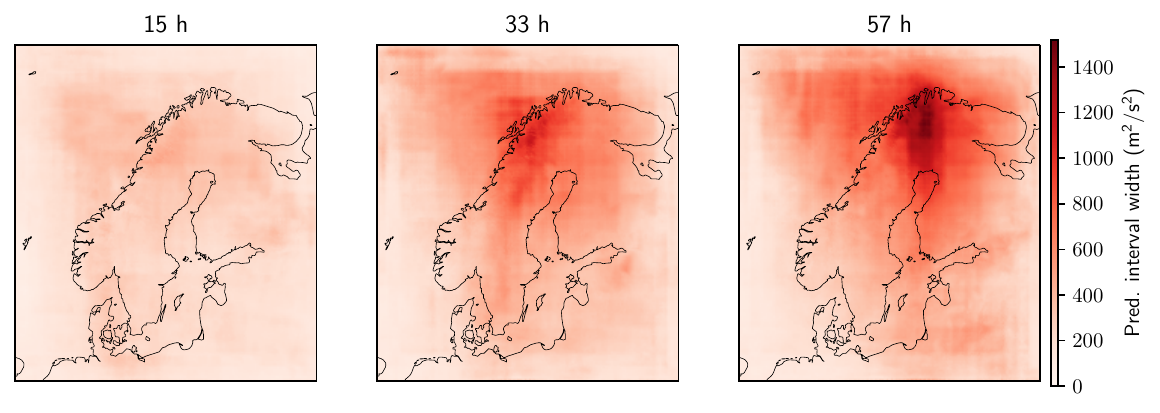}
        \caption*{\lammodelnll}
        \label{fig:neurwp_iwidth_z_nll}
    \end{subfloat}
    \caption{
    Width of predictive interval at $\alpha = 0.05$ for geopotential at 500 hPa (\texttt{z500}). 
    Both models here show a certain spatial pattern, which can be attributed to how the GNN in \lammodel is defined over the forecasting area.
    }
    \label{fig:neurwp_iwidth_z}
\end{figure}

\cref{fig:temp_2m} shows the ground truth, predicted forecast and the conformalised error intervals for temperature 2m above ground. 
Considering the autoregressive nature of \lammodel, the error accumulates and grows further in time, which is accurately captured by the CP framework (refer \cref{fig:neurwp_iwidth_z}).

We visualise the uncertainty for specific lead times by plotting the width of the error bars for all spatial locations in an example forecast.
Such plots for shortwave solar radiation are shown in \cref{fig:neurwp_iwidth_nswrs} and for geopotential in \cref{fig:neurwp_iwidth_z}.
In \cref{sec:appendix_results} we show plots for additional variables, including slice plots to help visualise the error bars obtained for different $\alpha$ values using the CP framework.

\Cref{fig:neurwp_iwidth_nswrs} highlights an important difference between the two methods for computing non-conformity scores. 
As the shortwave solar radiation is close to 0 during the night, it is easy for the model to predict.
During the day this is far more challenging, with the solar radiation being filtered through sparse and dynamic cloud cover.
With the RES non-conformity scores, used for \lammodelmse in \cref{fig:neurwp_iwidth_nswrs}, the width of the predictive intervals are determined during calibration, and does not change depending on the forecast from the model.
As a specific lead time can fall both during day and night, depending on the initialization time, CP will give large error bars also during the night.
This can be compared to the results for \lammodelnll in \cref{fig:neurwp_iwidth_nswrs}, using STD non-conformity scores.
In this case the bounds are very tight for lead times during the night (33~h and 57~h).
It can also be noted that for \lammodelnll at lead time 15 h we see clear spatial features appearing in the error bars themselves.
This corresponds to higher forecast uncertainty in areas of rapid change.
The forecast-dependent patterns for \lammodelnll thus have desirable properties, but this relies on having a model that outputs (potentially uncalibrated) standard-deviations.

We also evaluate the empirical coverage, the fraction of the data that lays within the conformalised error bars.
From \cref{eq:coverage_tensor} this should match the chosen value of \hbox{$1-\alpha$}.
For \hbox{$1-\alpha = 90\%$} we get coverage of $91.09\%$ for \lammodelmse and $91.08\%$ for \lammodelnll.
This can be compared to a coverage of $74.62\%$ for bounds computed directly from the uncalibrated Gaussians output by \lammodelnll.
In \cref{sec:appendix_results} we plot the empirical coverage for a range of values of $1-\alpha$.
Overall the STD method applied to \lammodelnll gives tighter bounds than RES for \lammodelmse (0.96 vs 1.13 average width, in standardized units).

\section{Discussion}

Due to the chaotic nature of the weather system, capturing uncertainty in weather forecasts has long been an important consideration both in research and operations. Ensemble forecasting requires a computational cost proportional to the number of ensemble members. In contrast, CP offers a cheap method to immediately quantify forecast uncertainty for a time, position, and variable of interest, and can be applied to existing deterministic models. The CP framework does come with several limitations such as marginal coverage, lack of predictive distribution and the need for exchangeability. 
Though extensive work is being done to overcome these limitations, CP in its current form does pose significant utility in providing statistically valid error bars.

\FloatBarrier
\newpage

\bibliography{bibliography}
\bibliographystyle{icml2024}

\newpage
\appendix
\onecolumn
\section{Additional Results}
\label{sec:appendix_results}
\begin{figure}[tbp]
    \centering
    \begin{subfigure}{0.5\textwidth}
        \centering
        \includegraphics[width=\linewidth]{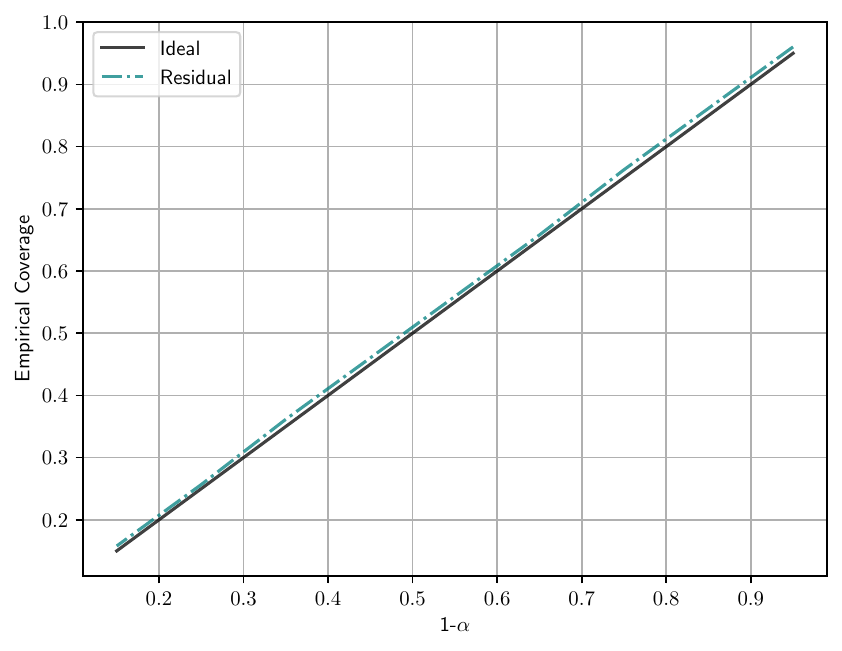}
        \caption{\lammodelmse}
    \end{subfigure}%
    \begin{subfigure}{0.5\textwidth}
        \centering
        \includegraphics[width=\linewidth]{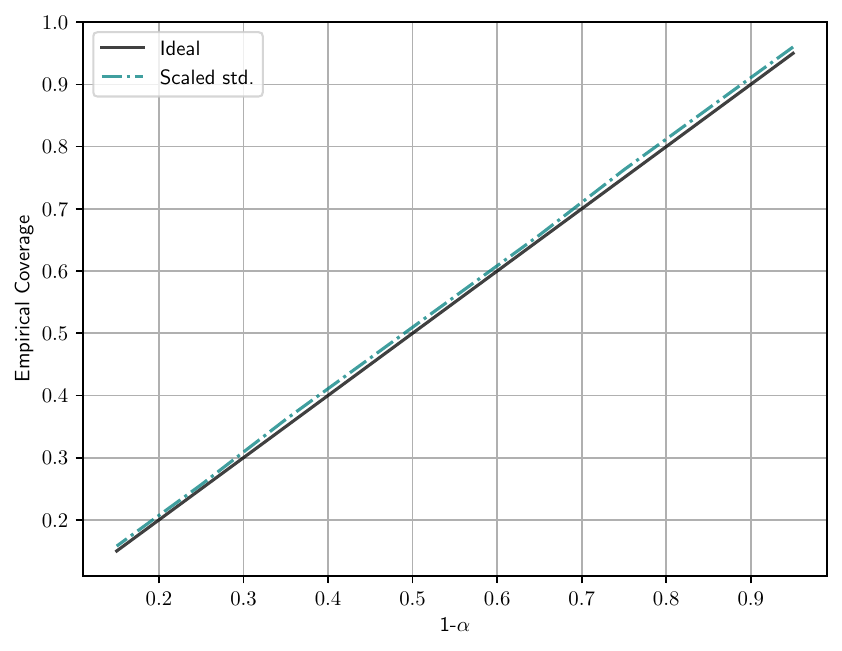}
        \caption{\lammodelnll}
    \end{subfigure}
    \caption{Empirical Coverage at different levels of $1-\alpha$ for \lammodelmse and \lammodelnll.}
    \label{fig:neurwp_emp_cov}
\end{figure}

\begin{figure}
    \centering
        \begin{subfigure}{0.5\textwidth}
        \centering
        \includegraphics[width=\linewidth]{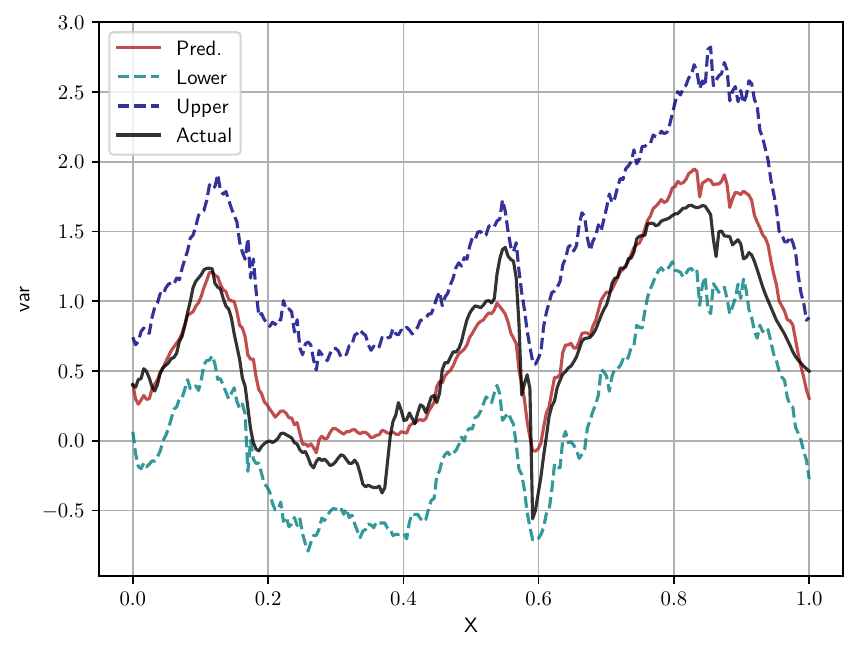}
        \caption{\lammodelmse, $\alpha = 0.05$}
        \end{subfigure}%
    \begin{subfigure}{0.5\textwidth}
        \centering
        \includegraphics[width=\linewidth]{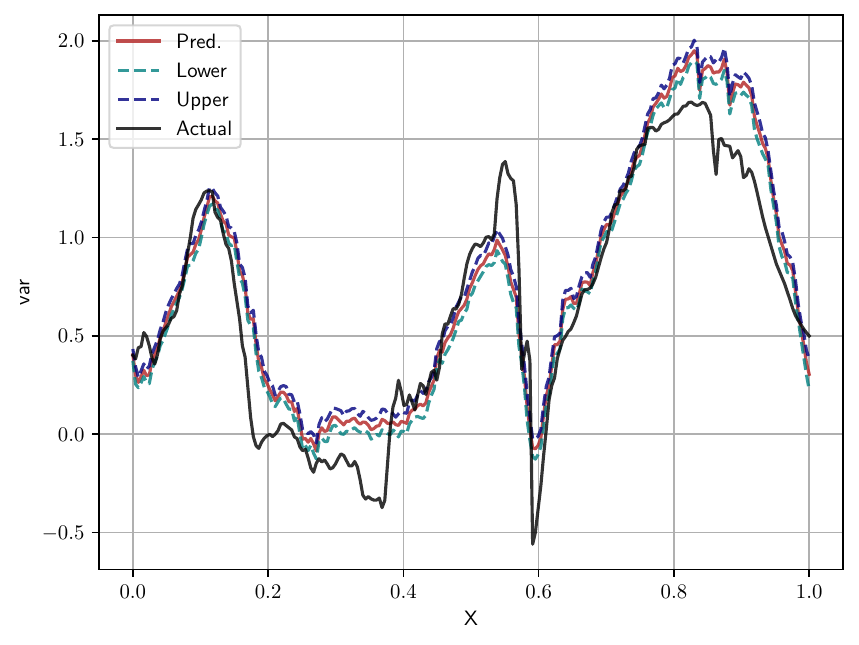}
        \caption{\lammodelmse, $\alpha = 0.85$}
    \end{subfigure}
    \begin{subfigure}{0.5\textwidth}
        \centering
        \includegraphics[width=\linewidth]{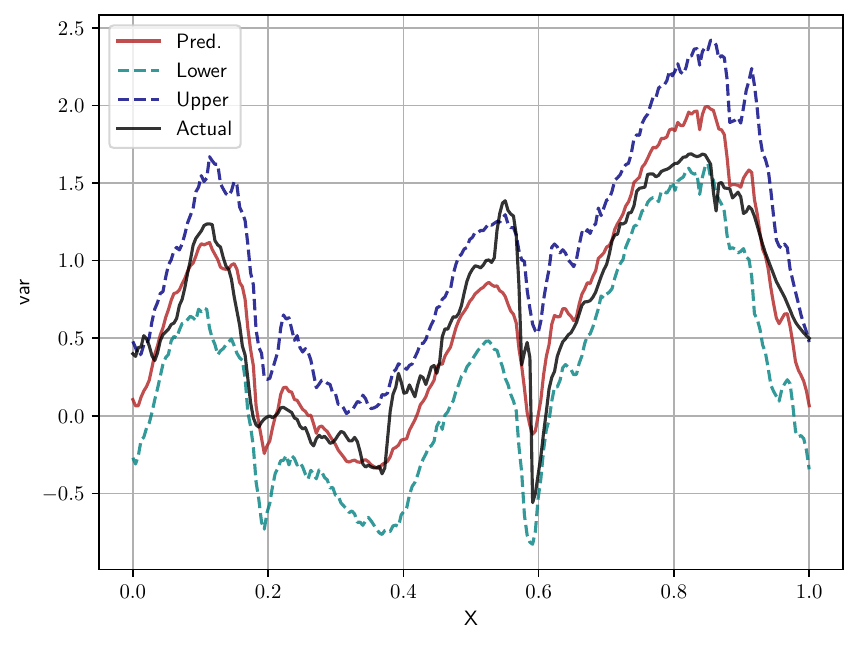}
        \caption{\lammodelnll, $\alpha = 0.05$}
        \end{subfigure}%
    \begin{subfigure}{0.5\textwidth}
        \centering
        \includegraphics[width=\linewidth]{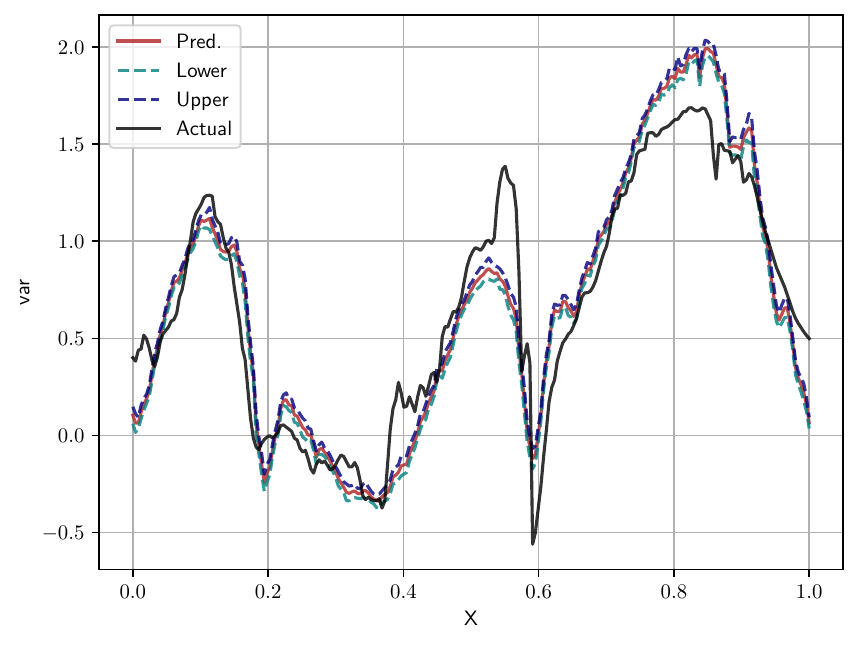}
        \caption{\lammodelnll, $\alpha = 0.85$}
    \end{subfigure}%
    \caption{Slice plots across the $x$-axis of a temporal prediction of a single variable (u-component of wind). Figures (a) - (d) depicts the ground truth, prediction, upper and lower bars obtained through the CP framework for the \lammodelmse and \lammodelnll for 95 percent coverage ($\alpha=0.05$) and 15 percent coverage ($\alpha=0.85$).}
    \label{fig:neurwp_slices}
\end{figure}

Figures \ref{fig:neurwp_iwidth_r_2} to \ref{fig:neurwp_iwidth_v_65} show examples of predictive intervals at $\alpha=0.05$ for additional variables from the models.
In figure \ref{fig:neurwp_emp_cov} we plot the empirical coverage for both models and different values of $1-\alpha$. In \cref{fig:neurwp_slices}, we show slice plots along the spatial domain to further highlight the error bars obtained at various coverage levels using the CP framework. 

\newcommand{\intervalfig}[2]{\begin{figure}[tbp]
    \centering
    \begin{subfloat}
        \centering
        \includegraphics[width=.7\linewidth]{Images/neurwp_plots/hi_lam_wmse/iwidth_alpha_0.05_#1.pdf}
        \caption*{\lammodelmse}
    \end{subfloat}
    \begin{subfloat}
        \centering
        \includegraphics[width=.7\linewidth]{Images/neurwp_plots/hi_lam_nll/iwidth_alpha_0.05_#1.pdf}
        \caption*{\lammodelnll}
    \end{subfloat}
    \caption{#2}
    \label{fig:neurwp_iwidth_#1}
\end{figure}}

\intervalfig{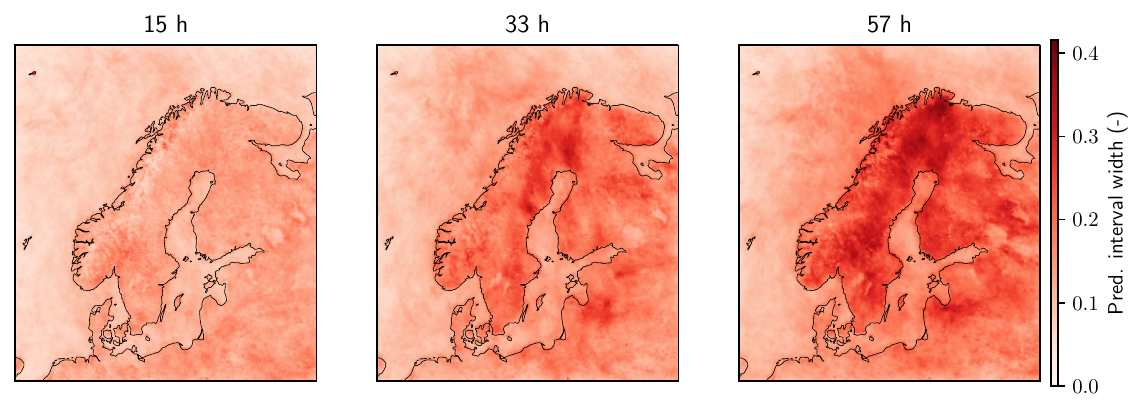}{
Width of predictive interval at $\alpha = 0.05$ for relative humidity at 2 m above ground (\texttt{2r}).
}
\intervalfig{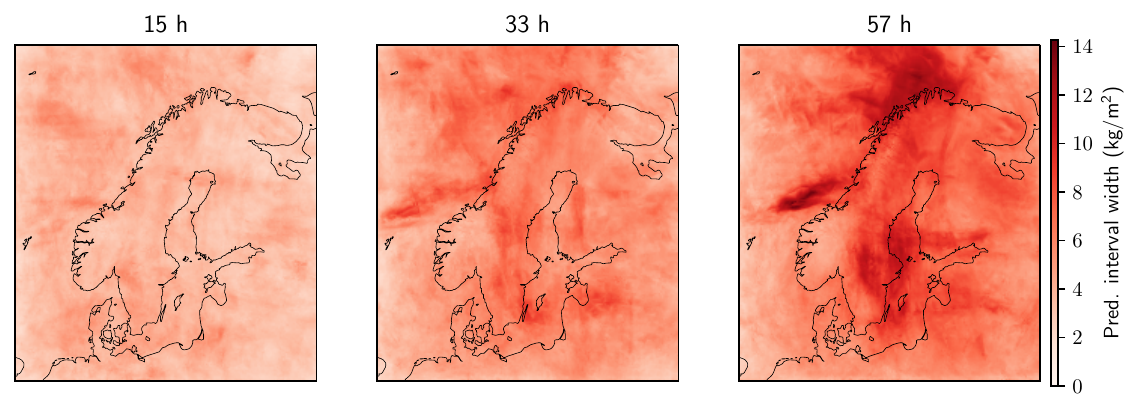}{
Width of predictive interval at $\alpha = 0.05$ for water vapour, integrated over the column above the grid cell (\texttt{wvint}).
}
\intervalfig{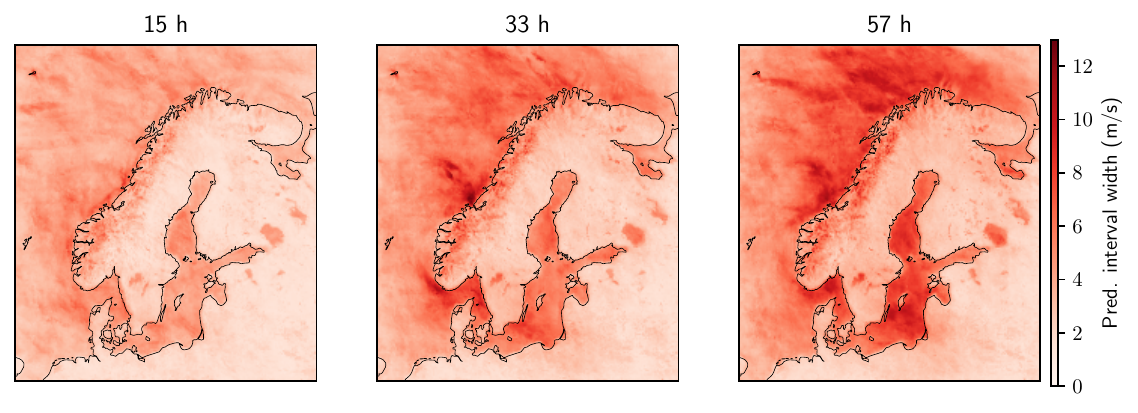}{
Width of predictive interval at $\alpha = 0.05$ for u-component of wind vector at model level 65 (\texttt{u65}).
Level 65 is the bottom level modelled in MEPS, and sits at approximately 12.5 m above ground.
}
\intervalfig{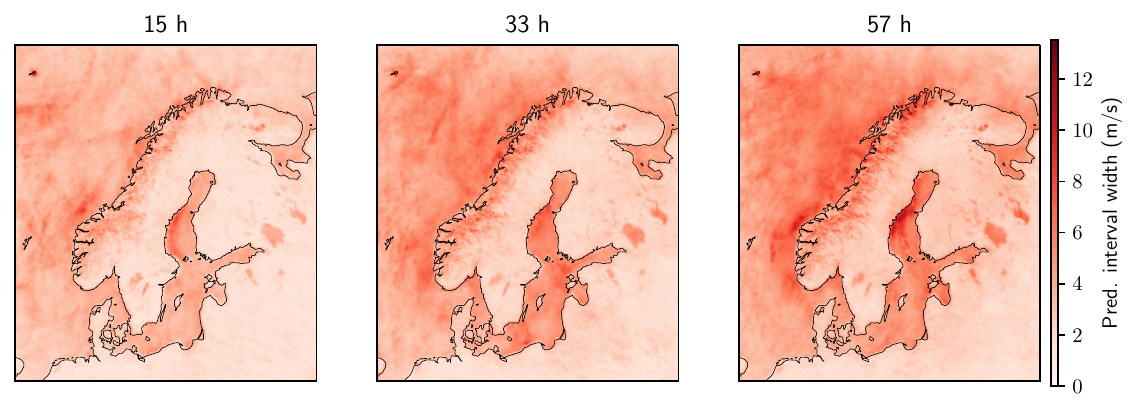}{
Width of predictive interval at $\alpha = 0.05$ for u-component of wind vector at model level 65 (\texttt{v65}).
Level 65 is the bottom level modelled in MEPS, and sits at approximately 12.5 m above ground.
}

\end{document}